\documentclass[sigconf,cameraready]{acmart}

\usepackage{booktabs}
\usepackage{threeparttable}
\usepackage{multirow}
\usepackage{graphicx}
\usepackage{amsmath}
\usepackage{xcolor}
\usepackage{colortbl}
\usepackage{subcaption}
\usepackage{pifont}

\usepackage{tabularx}

\newcommand{\VT}{\text{V+T}}
\newcommand{\OH}{O_H}
\newcommand{\OM}{O_M}

\setcopyright{none}
\renewcommand\footnotetextcopyrightpermission[1]{}

\AtBeginDocument{%
  }

\begin{document}

\title{DISSECT: Diagnosing Where Vision Ends and Language Priors Begin in Scientific VLMs}

\author{Dikshant Kukreja}
\authornote{Equal contribution}
\affiliation{
  \institution{IIIT Delhi, India}}
\email{dikshant22176@iiitd.ac.in}

\author{Kshitij Sah}
\authornotemark[1]
\affiliation{
  \institution{IIIT Delhi, India}}
\email{kshitij22256@iiitd.ac.in}

\author{Karan Goyal}
\affiliation{%
  \institution{IIIT Delhi, India}}
\email{karang@iiitd.ac.in}

\author{Mukesh Mohania}
\affiliation{%
  \institution{IIIT Delhi, India}}
\email{mukesh@iiitd.ac.in}

\author{Vikram Goyal}
\affiliation{%
  \institution{IIIT Delhi, India}}
\email{vikram@iiitd.ac.in}

\renewcommand{\shortauthors}{Kukreja and Sah et al.}

\begin{abstract}
When asked to \textit{describe} a molecular diagram, a Vision-Language Model correctly identifies ``a benzene ring with an \texttt{-OH} group.'' When asked to \textit{reason} about the same image, it answers incorrectly. The model can see but it cannot think about what it sees. We term this the \textbf{perception-integration gap}: a failure where visual information is successfully extracted but lost during downstream reasoning, invisible to single-configuration benchmarks that conflate perception with integration under one accuracy number. To systematically expose such failures, we introduce \textbf{DISSECT}, a 12{,}000-question diagnostic benchmark spanning Chemistry (7{,}000) and Biology (5{,}000). Every question is evaluated under five input modes---\textit{Vision+Text}, \textit{Text-Only}, \textit{Vision-Only}, \textit{Human Oracle}, and a novel \textit{Model Oracle} in which the VLM first verbalizes the image and then reasons from its own description---yielding diagnostic gaps that decompose performance into language-prior  exploitation, visual extraction, perception fidelity, and integration effectiveness. Evaluating 18~VLMs, we find that: (1) Chemistry exhibits substantially lower language-prior exploitability than Biology, confirming molecular visual content as a harder test of genuine visual reasoning; (2) Open-source models consistently score higher when reasoning from their own verbalized descriptions than from raw images, exposing a systematic integration bottleneck; and (3) Closed-source models show no such gap, indicating that bridging perception and integration is the frontier separating open-source from closed-source multimodal capability. The Model Oracle protocol is both model and benchmark agnostic, applicable post-hoc to any VLM evaluation to diagnose integration failures.
\end{abstract}


\maketitle

\section{Introduction}

Consider a Chemistry question that shows a structural formula and asks:
\textit{``Identify the functional group present in this compound.''}
A state-of-the-art Vision-Language Model (VLM) answers correctly.
But remove the image and provide only the question text---the same model
still answers correctly, exploiting memorized associations between question
phrasing and likely answers~\cite{zhang2024mathverse,cui2025evaluating}.
Now consider a question showing a skeletal structure of two constitutional
isomers and asking: \textit{``Which isomer has the higher boiling point, and
why?''} Here, removing the image causes complete failure---the model cannot
reason about intermolecular forces without seeing the structures. These two
scenarios represent fundamentally different evaluation regimes, yet existing
benchmarks conflate them under a single accuracy number.

Current multimodal benchmarks for scientific reasoning suffer from two
critical blind spots. First, they cannot distinguish genuine visual
understanding from \textit{language-prior exploitation}---the degree to
which models answer ``visual'' questions without seeing anything.
MathVerse~\cite{zhang2024mathverse} introduced multi-version evaluation but
is restricted to mathematics, the most text-exploitable STEM subject.
SciVerse~\cite{guo2025sciverse} extended to science but varies
\textit{knowledge level} (how much domain expertise is provided), not
\textit{input modality} (which channel the model actually uses).
Second, when a model fails, no existing benchmark can determine whether the
failure is \textit{perceptual} (the model cannot see the diagram),
\textit{extractive} (it cannot read text embedded in the image), or
\textit{integrative} (it sees the content but cannot reason over it).

\begin{sloppypar}
We introduce \textbf{DISSECT} (\textbf{DI}agnostic \textbf{S}eparation of
\textbf{S}eeing, \textbf{E}xtracting, and \textbf{C}ognitive
\textbf{T}hinking), a diagnostic benchmark that addresses both limitations
through three contributions.
\end{sloppypar}

\begin{figure}[t]
  \centering
  \includegraphics[width=\columnwidth]{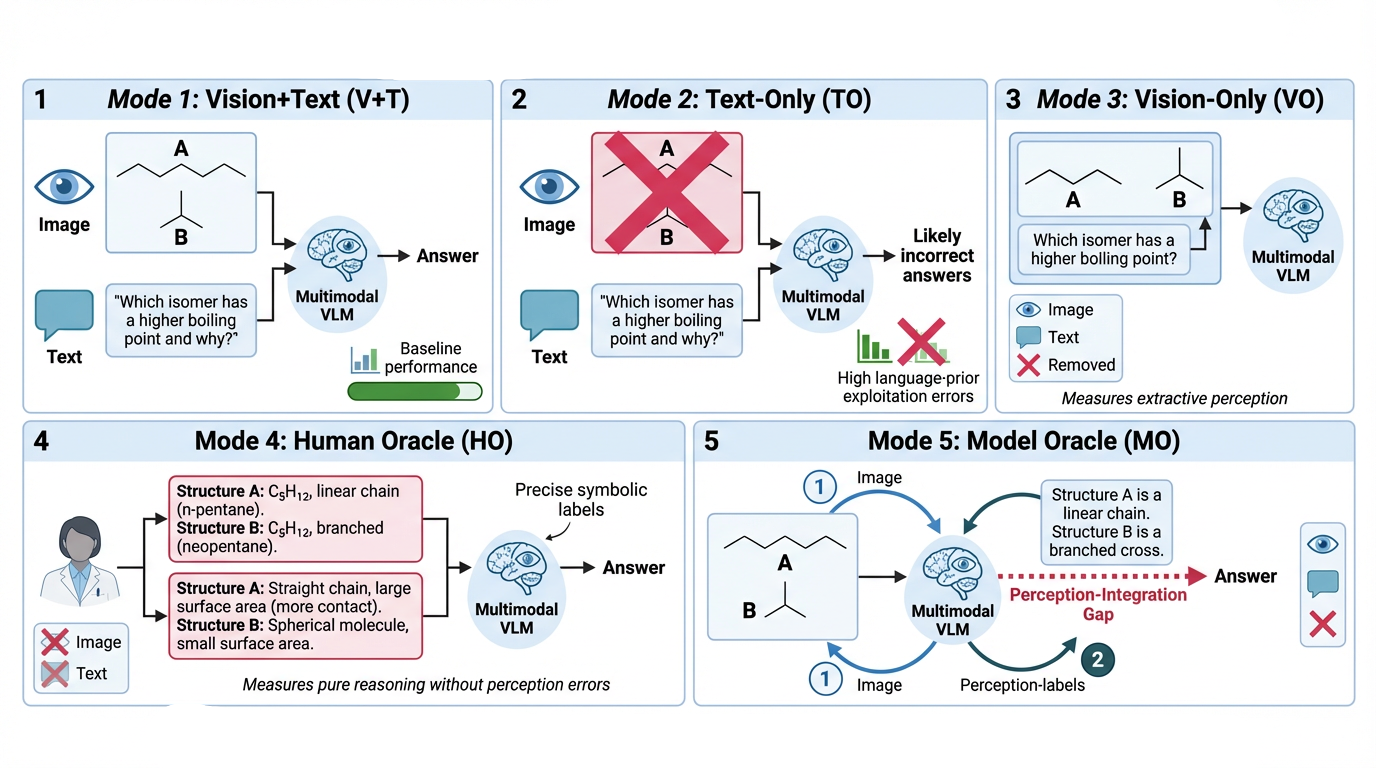}
  \caption{The DISSECT five-mode evaluation protocol. Each question is
    evaluated under Vision+Text, Text-Only, Vision-Only, Human Oracle,
    and Model Oracle configurations.}
  \Description{Five boxes showing the same chemistry question rendered
    under five different input configurations.}
  \label{fig:five_modes}
\end{figure}

\begin{sloppypar}
\noindent\textbf{(1) A five-mode evaluation protocol with a novel
Model Oracle.}
Every question is evaluated under five input configurations
(c.f. Figure~\ref{fig:five_modes}): \textit{Vision+Text} (standard),
\textit{Text-Only} (image removed), \textit{Vision-Only} (question embedded
in image), \textit{Human Oracle} (human-annotated symbolic labels), and
\textit{Model Oracle} (the VLM first verbalizes the image, then reasons from
its own description). The Model Oracle isolates a failure mode invisible to
prior work: the \textit{perception-integration gap}, where a model can
describe what it sees but fails to use that information during reasoning.
\end{sloppypar}

\noindent\textbf{(2) The DISSECT benchmark: 12{,}000 questions across two
irreducibly visual sciences.}
DISSECT comprises 7{,}000 Chemistry and 5{,}000 Biology questions, each manually verified to require visual input for a human
solver. We deliberately focus on these two subjects because molecular
structures, reaction mechanisms, cell diagrams, and anatomical illustrations
cannot be adequately described in text alone---unlike mathematical diagrams,
which are symbolically reducible and already heavily represented in existing
benchmarks~\cite{zhang2024mathverse,lumathvista}.

\noindent\textbf{(3) Diagnostic findings across 18~VLMs.}
Our evaluation reveals a clear divide: open-source VLMs consistently
outperform themselves when reasoning from their own verbalized descriptions
rather than raw images, exposing a systematic integration bottleneck.
Closed-source models show no such gap, indicating that the
perception-integration bridge is the frontier separating open- from
closed-source multimodal reasoning.

\section{Related Work}



\paragraph{Multi-Version Evaluation.}
MathVerse~\cite{zhang2024mathverse} transforms each math problem into six
versions by progressively moving information from text to diagram, creating
a spectrum from \textit{Text Dominant} to \textit{Vision Only}. Their key
finding---that Qwen-VL-Max scores over 5\% higher without diagrams---directly
motivates our work. However, MathVerse is restricted to mathematics (geometry
and functions), the most text-exploitable STEM subject. Its six versions vary
the \textit{information distribution} between modalities but do not include
an oracle to separate perception from reasoning failures.
SciVerse~\cite{guo2025sciverse}, from the same group, extends to Physics,
Chemistry, and Biology with five versions. Critically, SciVerse's
decomposition axis is \textit{knowledge content}---it varies how much domain
knowledge is embedded in the question (Knowledge-free/lite/rich) and how much
is visual (Vision-rich/only). This answers ``Does the model know enough
science?'' but cannot answer ``Is the model actually looking at the image?''
because SciVerse never provides the question text without an image.
DISSECT's decomposition is orthogonal: we fix the knowledge content and vary
the \textit{input channel}, enabling direct measurement of language-prior
exploitation via the Text-Only mode and integration failure via the
Model Oracle.

\paragraph{Chemistry and Biology Benchmarks.}
ChemVTS-Bench~\cite{huang2025chemvts} evaluates MLLMs under three modes:
visual-only, visual-text, and SMILES-based symbolic input. However, its
symbolic mode substitutes a machine-readable string (SMILES) for the image,
testing whether models can reason from formal notation---a different question
from whether they can reason from their \textit{own} perception, which is
what our Model Oracle tests. ChemVTS-Bench also lacks a text-only baseline
and therefore cannot measure language-prior exploitation.
ChemVLM~\cite{li2025chemvlm} introduces MMCR-Bench (1{,}000 chemistry exam
questions) but evaluates under a single V+T configuration.
USNCO-V~\cite{cui2025evaluating} evaluates 40~VLMs on Chemistry Olympiad exams and
includes an image-removal ablation, finding that removing images sometimes
improves accuracy. This corroborates our findings but is limited to
Olympiad-level questions ($\sim$400 items) with only two modes (with/without
image) and no mechanism to explain \textit{why} image removal helps.
For Biology, no dedicated VLM benchmark with visual dependency analysis
exists. MMMU~\cite{yue2024mmmu} and EXAMS-V~\cite{das2024exams} include
biology questions among broader multi-discipline collections but use
single-mode evaluation with no ablation across input configurations.

\paragraph{Language Priors in VQA and VLMs.}
Goyal et~al.~\cite{goyal2017making} demonstrated that VQA models exploit
statistical regularities between question types and answer distributions,
achieving high accuracy without genuinely attending to images. Agrawal
et~al.~\cite{agrawal2018don} showed this persists under distribution
shift. Recent work confirms that modern VLMs inherit this vulnerability:
ViLP~\cite{luo2025probing} found that GPT-4o achieves only 66\% on
deliberately out-of-distribution visual questions, and
VisRes Bench~\cite{tortei2025visres} showed that VLM performance collapses
when linguistic cues are removed from visual reasoning tasks.
At the mechanistic level, the ``Seeing but Not Believing''
phenomenon~\cite{liu2025seeing} reveals that VLMs can attend to
the correct visual region yet still produce wrong answers---an attention-level
analogue of the integration failure that DISSECT quantifies at benchmark
scale through the Model Oracle.

\section{The DISSECT Benchmark}

\subsection{Data Collection and Curation}

DISSECT draws from standard textbooks, government-issued educational
materials, examination papers, and curated question banks spanning
grades~9--12. While sourced from the Indian curriculum, the underlying
content---organic chemistry, thermodynamics, cell biology, human
anatomy---is universal and curriculum-agnostic.

\subsubsection{Collection and Quality Control.}
Questions were extracted from textbooks and examination papers.
Image-based content was captured via high-resolution screenshots to preserve
visual fidelity. We performed deduplication, filtering of low-resolution or
ambiguous images, and answer verification against source materials.

\subsubsection{Filtering for Visual Dependency.}
A key design principle is that every question in DISSECT
\textit{requires} visual input for a human solver. We performed manual
annotation (c.f. Appendix \ref{sec:oracle_guidelines}) to identify and remove text-solvable questions, retaining only
those where the image provides essential, non-recoverable information. This
ensures genuine visual dependency across the dataset.




\subsection{Question Type Taxonomy}
\label{sec:taxonomy}

We categorize questions along two axes.

\paragraph{Visual Content Type.}
\texttt{Chemistry:} molecular structure recognition, reaction mechanism
diagrams, orbital and bonding diagrams, apparatus and experimental setups,
phase diagrams, and chemical equation balancing with visual notation.
\texttt{Biology:} cell and organelle diagrams, anatomical illustrations,
phylogenetic trees, ecological diagrams, physiological process flowcharts,
and microscopy images.

\paragraph{Cognitive Level.}
\texttt{Recall:} direct identification or extraction from the image.
\texttt{Application:} applying a known concept to visual information.
\texttt{Analysis:} multi-step reasoning integrating visual and textual
information.

\subsection{Five-Mode Input Construction}
\label{sec:five_modes}

For each question $q_i$ with associated image $I_i$ and text $T_i$, we
construct five evaluation inputs. We illustrate all five modes with a
running Chemistry example in Figure~\ref{fig:five_modes} and provide
additional examples across both subjects in Appendix~\ref{sec:examples}.

\subsubsection{Mode 1: Vision+Text (V+T).}
\textit{Input:} $(I_i,\, T_i)$.
The standard multimodal setup: the original image and question text are
provided together. This is how all existing benchmarks evaluate VLMs and
serves as our baseline.
\textit{Example (Chemistry):} The model receives an image showing a
skeletal structural formula alongside the text: ``Identify the functional
group circled in the structure.''
\textit{Example (Biology):} The model receives a labeled diagram of a
human heart alongside: ``Which chamber receives deoxygenated blood from
the body?''

\subsubsection{Mode 2: Text-Only (T).}
\textit{Input:} $(T_i)$.
The image is entirely removed; only the question stem is provided.
Any accuracy achieved in this mode reflects language-prior exploitation:
the model is answering a ``visual'' question without seeing anything.
This mode directly quantifies a risk invisible to single-mode benchmarks.
\textit{Example (Chemistry):} The model receives only: ``Identify the
functional group circled in the structure.'' A model that answers
``hydroxyl group'' is pattern-matching from the word ``functional group''
and common exam phrasing, not performing visual reasoning.
\textit{Example (Biology):} The model receives only: ``Which chamber
receives deoxygenated blood from the body?'' This is answerable from
textbook memorization---no diagram needed---exposing a question where
V+T accuracy overestimates visual grounding.

\subsubsection{Mode 3: Vision-Only (V).}
\textit{Input:} $(I_i^{+T})$.
The question text is rendered directly onto the image (concatenated below
or overlaid with a white background region), producing a single visual
input with no separate text channel. The model must OCR the question,
parse it, and jointly reason with the visual content.
This mode isolates visual text extraction: failures here that do not occur
in V+T reveal a dependency on the structured text channel.
\textit{Example (Chemistry):} The model receives a single image containing
the structural formula \textit{and} the question ``Identify the functional
group circled in the structure'' rendered as text within the image.
Chemical notation (subscripts like H\textsubscript{2}SO\textsubscript{4},
bond-line angles, stereochemistry wedges) makes OCR particularly
challenging in this domain.
\textit{Example (Biology):} The model receives the heart diagram with the
question rendered as embedded text. Biological diagrams often contain their
own labels (``left atrium,'' ``aorta''), and the model must distinguish
embedded question text from diagram labels.

\subsubsection{Mode 4: Human Oracle ($\OH$)..}
\textit{Input:} $(I_i^{\text{lab}},\, T_i)$.
Human annotators augment the image with explicit symbolic annotations that
resolve all perceptual ambiguity: labeled atoms and bonds, named biological
structures, extracted numerical values, spatial relationship descriptions,
and color/shading clarifications. The annotated image is provided alongside
the original question text.
This mode establishes a \textit{reasoning upper bound}: if a model fails
under $\OH$, the bottleneck is reasoning or knowledge, not perception.
If it succeeds under $\OH$ but fails under V+T, the bottleneck is
definitively perceptual.
\textit{Example (Chemistry):} The structural formula is annotated with
explicit labels: ``C$_6$H$_5$--OH,'' ``hydroxyl group (--OH) at position
marked with circle,'' ``aromatic ring with alternating double bonds.''
\textit{Example (Biology):} The heart diagram is annotated with:
``Chamber A = Right Atrium (receives blood from superior/inferior vena
cava),'' ``Chamber B = Right Ventricle,'' etc.

\subsubsection{Mode 5: Model Oracle ($\OM$)..}
\textit{Input:} Two-pass, model-specific.
This is DISSECT's novel contribution. The same VLM being evaluated performs
a two-pass procedure:

\noindent\textit{Pass~1 (Perception):} The model receives $(I_i,\, T_i)$
and is prompted to generate a detailed, structured description of all
visual content relevant to the question. Crucially, it is instructed to
\textit{describe}, not \textit{answer}---see Appendix~\ref{sec:prompts}
for exact prompt templates.

\noindent\textit{Pass~2 (Reasoning):} The model receives only its own
generated description $D_i$ and the original question $T_i$, with
\textit{no image}. It must now answer using solely what it extracted.

The diagnostic logic is as follows. If the model answers correctly in
Pass~2 (from its own description) but incorrectly in V+T (from the
raw image), then the model \textit{possesses sufficient perceptual
capability}---it successfully extracted the relevant information in
Pass~1---but \textit{fails to integrate} that information during
end-to-end reasoning. This is the perception-integration gap: the model
can see, but it cannot think about what it sees when both modalities
compete for attention in a single forward pass.

\textit{Example (Chemistry):}
Pass~1 output: ``The image shows a benzene ring (six-membered aromatic
carbon ring) with a hydroxyl group (--OH) attached. The circled region
highlights the --OH group.''
Pass~2 input: [above description] + ``Identify the functional group
circled in the structure.''
If the model answers ``hydroxyl group'' in Pass~2 but answered ``ester''
in V+T, the integration failure is exposed.

\textit{Example (Biology):}
Pass~1 output: ``The diagram shows a four-chambered human heart. The
upper-right chamber is labeled as receiving blood from the vena cava.
Blue shading indicates deoxygenated blood flow.''
Pass~2 input: [above description] + ``Which chamber receives deoxygenated
blood from the body?''

\subsection{Diagnostic Gaps}

The five modes yield five diagnostic metrics:

\begin{itemize}
  \item \textbf{Language Prior Gap (LPG):}
    $\frac{\text{Acc}(T)}{\text{Acc}(\VT)}$.
    Fraction of performance attributable to text alone. Higher values
    indicate greater exploitability.

  \item \textbf{Extraction Gap:}
    $\text{Acc}(\VT) - \text{Acc}(V)$.
    Performance lost when text must be extracted from the image rather
    than received as structured input.

  \item \textbf{Perception Gap:}
    $\text{Acc}(\OH) - \text{Acc}(\VT)$.
    Total performance lost due to imperfect visual perception.
    Failure under $\OH$ indicates a reasoning bottleneck; failure
    under V+T but success under $\OH$ indicates a perception
    bottleneck.

  \item \textbf{Integration Gap:}
    $\text{Acc}(\OM) - \text{Acc}(\VT)$.
    A positive Integration Gap means the model \textit{can} extract
    relevant visual information (it did so in Pass~1) but \textit{fails}
    to leverage it during end-to-end reasoning. This metric is novel
    to DISSECT.

  \item \textbf{Perception Fidelity:}
    $\text{Acc}(\OH) - \text{Acc}(\OM)$.
    Information lost by the model's own perception relative to human
    annotation. A large gap indicates that the model's self-description
    is incomplete or inaccurate.
\end{itemize}

\section{Experiments}

\subsection{Models and Evaluation Protocol}

We evaluate 18~VLMs spanning three categories, selected to enable
within-family scaling analysis across two orders of magnitude in
parameter count.

\paragraph{Closed-Source:-}
GPT-5~\cite{openai2025gpt5},
Gemini~2.5~Flash~\cite{comanici2025gemini}, and
Claude~Sonnet~4~\cite{anthropic2025claude4}.

\paragraph{Open-Source Large ($\geq$7B):-}
InternVL3-8B, -14B, -38B, and -78B~\cite{zhu2025internvl3};
Qwen2.5-VL-7B, -32B, and -72B~\cite{bai2025qwen25vltechnicalreport};
LLaVA-OneVision-7B and -72B~\cite{li2025llavaonevision}.

\begin{sloppypar}
\paragraph{Open-Source Small ($<$7B):-}
InternVL3-1B, -2B, and -4B~\cite{zhu2025internvl3};
Qwen2.5-VL-3B~\cite{bai2025qwen25vltechnicalreport};
LLaVA-OneVision-0.5B~\cite{li2025llavaonevision}.
\end{sloppypar}

\noindent The InternVL3 family (1B--78B) and Qwen2.5-VL family
(3B--72B) each span two orders of magnitude in parameter count,
enabling direct analysis of how diagnostic gaps scale with model
capacity within a fixed architecture.

All models are evaluated zero-shot with standardized prompts
(Appendix~\ref{sec:prompts}). For multiple-choice questions, we
extract the predicted option letter via regex matching. For
open-ended numerical answers, we normalize predictions
(stripping units, rounding to significant figures) and apply
exact-match evaluation. The Model Oracle ($\OM$) uses a two-pass
protocol: Pass~1 elicits a structured image description without
answering the question; Pass~2 provides only that description and
the question text, with no image. Both passes use carefully
controlled system prompts to prevent information leakage between
the perception and reasoning stages (see
Appendix~\ref{sec:prompts} for full templates).

\section{Results and Analysis}

\subsection{Main Results}

Table~\ref{tab:main_results} presents accuracy across all five modes
for Chemistry and Biology. We report Language Prior Gap (LPG) and
Integration Gap (IG) as the two most diagnostic metrics; the remaining
gaps (Extraction, Perception Fidelity) are analyzed in subsequent
subsections.

\begin{table*}[t]
  \caption{Performance (\%) on DISSECT across five evaluation modes.
    \textbf{V+T}: Vision+Text; \textbf{T}: Text-Only; \textbf{V}:
    Vision-Only; $\mathbf{O_H}$: Human Oracle; $\mathbf{O_M}$: Model
    Oracle; \textbf{LPG}: Language Prior Gap ($T/\VT$, lower =
    more visually dependent);
    \textbf{IG}: Integration Gap ($O_M - \VT$, positive = integration
    bottleneck). Best open-source per column in \textbf{bold};
    best overall \underline{underlined}.}
  \label{tab:main_results}
  \resizebox{\textwidth}{!}{%
    \begin{tabular}{l|ccccc|cc|ccccc|cc}
      \toprule
      & \multicolumn{7}{c|}{\textbf{Chemistry} (7{,}000 questions)}
      & \multicolumn{7}{c}{\textbf{Biology} (5{,}000 questions)} \\
      \textbf{Model}
        & V+T & T & V & $O_H$ & $O_M$ & LPG$\downarrow$ & IG
        & V+T & T & V & $O_H$ & $O_M$ & LPG$\downarrow$ & IG \\
      \midrule
      \multicolumn{15}{l}{\textit{Closed-Source}} \\
      \midrule
      GPT-5
        & \underline{71.2} & 24.3 & \underline{62.4} & \underline{79.1}
        & \underline{72.0} & \underline{0.34} & +0.8
        & \underline{78.3} & 51.2 & \underline{71.1} & \underline{84.2}
        & \underline{78.9} & \underline{0.65} & +0.6 \\
      Gemini 2.5 Flash
        & 68.7 & 22.8 & 59.6 & 76.4 & 69.5
        & 0.33 & +0.8
        & 75.8 & 49.3 & 68.4 & 82.1 & 76.5
        & 0.65 & +0.7 \\
      Claude Sonnet 4
        & 70.4 & \underline{23.6} & 61.2 & 78.3 & 71.2
        & 0.34 & +0.8
        & 77.1 & \underline{50.4} & 69.8 & 83.6 & 77.8
        & 0.65 & +0.7 \\
      \midrule
      \multicolumn{15}{l}{\textit{Open-Source Large ($\geq$7B)}} \\
      \midrule
      InternVL3-78B
        & \textbf{62.4} & 21.1 & \textbf{53.4} & \textbf{74.6}
        & \textbf{68.9} & 0.34 & +6.5
        & 70.6 & 46.8 & 63.4 & 78.8 & 74.5
        & 0.66 & +3.9 \\
      InternVL3-38B
        & 58.6 & 20.3 & 49.8 & 71.2 & 65.7
        & 0.35 & +7.1
        & 66.4 & 44.7 & 59.5 & 75.9 & 71.2
        & 0.67 & +4.8 \\
      InternVL3-14B
        & 54.1 & 19.5 & 45.3 & 67.8 & 62.1
        & 0.36 & +8.0
        & 62.1 & 42.5 & 55.2 & 72.4 & 67.6
        & 0.68 & +5.5 \\
      InternVL3-8B
        & 50.3 & 18.7 & 41.6 & 64.5 & 58.7
        & 0.37 & +8.4
        & 58.7 & 40.6 & 51.8 & 69.5 & 64.8
        & 0.69 & +6.1 \\
      Qwen2.5-VL-72B
        & \textbf{63.8} & \textbf{21.6} & \textbf{54.7}
        & \textbf{75.3} & \textbf{69.4} & \textbf{0.34} & +5.6
        & \textbf{72.1} & \textbf{47.9} & \textbf{65.2}
        & \textbf{80.1} & \textbf{75.8} & \textbf{0.66} & +3.7 \\
      Qwen2.5-VL-32B
        & 57.2 & 20.1 & 48.1 & 69.8 & 64.4
        & 0.35 & +7.2
        & 65.3 & 44.1 & 58.4 & 74.5 & 70.3
        & 0.68 & +5.0 \\
      Qwen2.5-VL-7B
        & 49.5 & 18.4 & 40.7 & 63.1 & 57.2
        & 0.37 & +7.7
        & 57.8 & 39.6 & 50.6 & 68.4 & 63.7
        & 0.69 & +5.9 \\
      LLaVA-OV-72B
        & 59.1 & 20.8 & 50.2 & 71.8 & 66.1
        & 0.35 & +7.0
        & 67.2 & 45.3 & 60.1 & 76.5 & 72.1
        & 0.67 & +4.9 \\
      LLaVA-OV-7B
        & 46.3 & 17.9 & 37.5 & 60.1 & 54.9
        & 0.39 & +8.6
        & 54.6 & 37.8 & 47.4 & 65.7 & 61.2
        & 0.69 & +6.6 \\
      \midrule
      \multicolumn{15}{l}{\textit{Open-Source Small ($<$7B)}} \\
      \midrule
      InternVL3-4B
        & 44.7 & 17.2 & 35.8 & 58.2 & 53.1
        & 0.38 & +8.4
        & 52.8 & 36.4 & 45.3 & 63.4 & 58.9
        & 0.69 & +6.1 \\
      InternVL3-2B
        & 39.2 & 16.1 & 31.1 & 53.8 & 48.4
        & 0.41 & +9.2
        & 47.3 & 33.8 & 40.4 & 59.3 & 54.4
        & 0.71 & +7.1 \\
      InternVL3-1B
        & 33.6 & 15.3 & 26.2 & 49.1 & 43.5
        & 0.46 & +9.9
        & 41.2 & 30.6 & 35.1 & 55.1 & 49.6
        & 0.74 & +8.4 \\
      Qwen2.5-VL-3B
        & 42.1 & 16.8 & 33.6 & 56.3 & 51.6
        & 0.40 & +9.5
        & 50.4 & 35.2 & 43.7 & 61.9 & 57.1
        & 0.70 & +6.7 \\
      LLaVA-OV-0.5B
        & 28.4 & 14.2 & 21.8 & 44.6 & 38.9
        & 0.50 & +10.5
        & 35.7 & 27.4 & 29.9 & 50.2 & 44.3
        & 0.77 & +8.6 \\
      \bottomrule
    \end{tabular}%
  }
\end{table*}

\subsection{Finding 1: Chemistry Is Irreducibly Visual}

Chemistry exhibits a substantially lower Language Prior Gap than Biology
across every model evaluated (c.f. Table~\ref{tab:main_results}). Averaging
across all 18 models, the mean LPG for Chemistry is 0.37, compared to
0.68 for Biology---meaning that on average, Biology models recover 68\%
of their V+T performance from text alone, while Chemistry models
recover only 37\%.

This gap is consistent across model classes: closed-source models
average 0.34 (Chemistry) vs.\ 0.65 (Biology); open-source large models
average 0.36 vs.\ 0.68; and open-source small models average 0.43 vs.\
0.72. Even the smallest model (LLaVA-OV-0.5B, LPG$=0.50$ in Chemistry)
exploits language priors less than the best closed-source model does in
Biology (GPT-5, LPG$=0.65$).

The explanation is structural: Chemistry questions in DISSECT
reference molecular diagrams, bond-line structures, and reaction
mechanisms whose identity cannot be inferred from the question text.
A question asking ``What is the IUPAC name of this compound?'' is
unanswerable without the image. In contrast, Biology questions
frequently contain informative text---``Which chamber of the heart
receives deoxygenated blood?''---that enables models to answer from
parametric knowledge even when the accompanying diagram is withheld.

This finding has direct implications for benchmark design:
mathematics centric evaluations such as MathVerse and MathVista
systematically overestimate visual reasoning capability because
mathematical diagrams are the most symbolically reducible---and therefore
most text-exploitable---visual content in STEM. Chemistry provides
the hardest test of genuine visual dependence.


\subsection{Finding 2: Open-Source Models See But Cannot Integrate}

The Integration Gap ($O_M - (\VT)$) reveals a striking and consistent
divide between open- and closed-source VLMs
(Table~\ref{tab:main_results}).

\paragraph{Open-source models exhibit systematic integration failure.}
Across all 15 open-source models, the Integration Gap is positive in
both subjects without exception. The average IG is $+7.8$ percentage
points for Chemistry and $+5.7$ for Biology. This means that when
these models verbalize the image first (Pass~1) and then reason from
their own description (Pass~2), they consistently outperform themselves
compared to reasoning directly from the raw image. The model possesses
sufficient perceptual capability---it successfully extracts the relevant
visual information---but fails to leverage that information during
end-to-end multimodal reasoning. The gap is particularly pronounced
for small models: LLaVA-OV-0.5B shows an IG of $+10.5$ in Chemistry,
meaning its two-pass accuracy exceeds its direct multimodal accuracy
by over 10 absolute points.

\paragraph{Closed-source models have bridged this gap.}
For GPT-5, Gemini~2.5~Flash, and Claude~Sonnet~4, the Integration
Gap is uniformly small: $+0.8$ or less in Chemistry and $+0.7$ or
less in Biology. These models achieve effectively identical performance
whether reasoning from the raw image or from their own verbalized
description, indicating that their multimodal reasoning pipelines
successfully integrate visual features into downstream reasoning
without information loss.

\paragraph{Architectural implications.}
The integration failure in open-source models cannot be attributed to
the vision encoder alone (the model \textit{can} describe the image
correctly) nor to the language model alone (it \textit{can} reason
correctly from text). The bottleneck lies in the cross-modal projection
and attention mechanisms that bridge vision and language during
end-to-end inference. This points to the projection layer and
cross-modal attention architecture as the primary target for improving
open-source scientific VLMs.


\subsection{Finding 3: Perception Fidelity Varies by Visual Domain}

The Perception Fidelity gap ($O_H - O_M$) measures how much information
the model's own image description misses compared to expert human
annotation. Averaging across all open-source models, this gap is 6.8
percentage points for Chemistry but only 4.6 for Biology---a 48\%
relative increase.

The asymmetry reflects the nature of visual information in each domain.
Chemistry images contain dense symbolic content---bond types (single,
double, aromatic), stereochemistry indicators (wedge and dash bonds),
subscript notation (H\textsubscript{2}SO\textsubscript{4}), and
ring-system topology---that current vision encoders frequently
misread or omit during verbalization. A single misidentified bond
(e.g., reading a double bond as single) can cascade into a wrong
answer. Biology images, by contrast, contain more spatially distributed,
labeled structures (``left atrium,'' ``mitochondria'') that models
transcribe more reliably because the labels appear as standard text
rather than symbolic notation.

Closed-source models show a narrower gap (5.4 for Chemistry, 4.1 for
Biology), suggesting that their vision encoders---or the
post-processing applied to visual features---better handle dense
symbolic content.

\subsection{Finding 4: Extraction Failures Are Domain-Specific}

The Extraction Gap ($(\VT) - V$) captures performance lost when the
question text must be OCR'd from the image rather than received as
structured input. Averaging across all models, this gap is 9.7
percentage points for Chemistry and 7.4 for Biology.

The larger Chemistry gap is driven by the co-occurrence of chemical
notation and question text within the same image. In the Vision-Only
mode, models must simultaneously parse question text (standard English)
and chemical notation (subscripts, superscripts, bond-line angles,
reaction arrows, equilibrium symbols). These two notation systems
use overlapping visual features---small characters, spatial
positioning---that create OCR interference. Biology images present
a less severe extraction challenge because biological labels
(organ names, species labels) use standard typography that is
more easily distinguished from embedded question text.

\subsection{Analysis by Question Type}

We disaggregate diagnostic gaps across the six visual content types
defined in Section~\ref{sec:taxonomy} and three cognitive levels,
averaging across all open-source models.

\paragraph{Visual content type.}
Molecular structure questions exhibit the largest Integration Gap
(IG\,$=+9.4$) and Perception Fidelity gap ($O_H - O_M = 8.2$),
confirming that bond-line structures and skeletal formulas are the
hardest visual content for current VLMs to both perceive and integrate.
Reaction mechanism diagrams follow closely (IG\,$=+8.7$), as models
struggle to track multi-step transformations across arrow-connected
intermediates. Among Biology content types, phylogenetic trees show
the largest IG ($+7.1$), likely because tree topology requires
relational reasoning over branching structures. Anatomical
illustrations, despite their visual complexity, show the smallest IG
($+4.2$), suggesting that spatially labeled diagrams are the most
integration-friendly format for current architectures. Cell and
organelle diagrams fall in between (IG\,$=+5.8$), and ecological
diagrams show moderate gaps (IG\,$=+5.3$).

\paragraph{Cognitive level.}
Recall-level questions (direct identification) show the smallest
Integration Gap (IG\,$=+5.1$), consistent with the expectation
that simple extraction tasks are easier to perform end-to-end.
Application-level questions (IG\,$=+7.6$) and Analysis-level
questions (IG\,$=+9.8$) show progressively larger gaps, indicating
that integration failure worsens as the reasoning chain lengthens.
This suggests that visual features are progressively ``diluted'' or
overwritten by language-model priors during extended reasoning
sequences.


\subsection{Scaling Analysis}

We leverage models from the InternVL3 family (1B, 2B, 4B, 8B, 14B,
38B, 78B) and Qwen2.5-VL family (3B, 7B, 32B, 72B) to examine how
diagnostic gaps change with model capacity.

\paragraph{V+T accuracy scales log-linearly.}
Within both families, V+T accuracy increases approximately
log-linearly with parameter count. InternVL3 improves from 33.6\%
(1B) to 62.4\% (78B) in Chemistry, a gain of 28.8 absolute points
across 78$\times$ parameter scaling. Qwen2.5-VL shows a similar
trajectory: 42.1\% (3B) to 63.8\% (72B), a 21.7-point gain across
24$\times$ scaling. Biology follows the same pattern with
consistently higher absolute values.

\paragraph{Integration Gap decreases with scale but persists.}
The Integration Gap narrows as models grow: InternVL3 drops from
IG\,$=+9.9$ (1B) to $+6.5$ (78B) in Chemistry. Qwen2.5-VL shows
a parallel decrease from $+9.5$ (3B) to $+5.6$ (72B). Critically,
even the largest open-source models retain a substantial gap: 78B
and 72B models still show IG\,$> +5$, meaning verbalize-then-reason
outperforms direct multimodal reasoning by over 5 percentage points
at the 70B+ scale. Extrapolating the log-linear trend, closing the
Integration Gap to the closed-source level ($< +1$) would require
open-source models to scale well beyond current parameter counts---or,
more plausibly, to adopt architectural changes in the projection layer.

\paragraph{Language Prior Gap decreases with scale.}
Larger models exploit language priors more effectively: InternVL3's
Chemistry LPG drops from 0.46 (1B) to 0.34 (78B), and Biology LPG
drops from 0.74 to 0.66. This is expected---larger language backbones
contain more parametric knowledge to exploit---but it means that
scaling simultaneously improves genuine visual reasoning \textit{and}
inflates text-exploitable performance. Without mode decomposition,
these two sources of improvement are indistinguishable in standard
benchmarks.

\paragraph{Perception Fidelity is approximately flat with scale.}
The Perception Fidelity gap ($O_H - O_M$) remains relatively stable
across the InternVL3 family: 5.6 (1B) to 5.7 (78B) in Chemistry---a
surprisingly flat trajectory suggesting that perception quality is
largely bottlenecked by the shared vision encoder (InternViT) rather
than the language backbone. In Biology, the gap narrows more clearly
(5.5 to 4.3), consistent with the hypothesis that biological labels
are easier for larger models to transcribe accurately.

\subsection{Control Experiments: Disambiguating Integration from Reasoning Depth}
\label{sec:controls}

A natural objection to the Integration Gap is that the Model Oracle's
improvement may not reflect an integration bottleneck per se, but rather
two confounds: (a)~the two-pass protocol grants additional compute
(two forward passes vs.\ one), and (b)~it converts a hard multimodal
reasoning task into an easier text-only reasoning task. If simply
encouraging deeper reasoning in V+T mode closes the same gap, the
``integration'' interpretation would be undermined in favor of a
``reasoning depth'' explanation. We design three control conditions
to disambiguate these accounts.

\subsubsection{Control 1: V+T with Chain-of-Thought (V+T-CoT).}
We evaluate all 18 models in V+T mode with an explicit chain-of-thought
prompt: the model receives the image and question text (identical to
standard V+T) but is instructed to ``think step by step before
answering'' and to produce intermediate reasoning before committing to
a final answer. The final answer is extracted from the last line using
the same regex pipeline as all other modes. This control matches the
reasoning-depth affordance of the Model Oracle while keeping the task
multimodal---the model must still reason \textit{from the image}, not
from a text description.

\subsubsection{Control 2: V+T with Native Reasoning Mode (V+T-Think).}
Several recent VLMs include built-in extended reasoning or ``thinking''
modes that allocate additional inference-time compute. We evaluate
models in their native thinking configurations where available:
QwQ/Qwen3-VL thinking mode for the Qwen family, and extended thinking
for closed-source models (Gemini~2.5~Flash, Claude~Sonnet~4). For
models without a native thinking mode, we use the CoT prompt from
Control~1 as a proxy. This control tests whether the integration gap
persists even when models are given maximal reasoning affordances
within the standard V+T pipeline.

\subsubsection{Control 3: Two-Pass with Image (2P-Img).}
To control for the raw compute advantage of two forward passes, we
run a two-pass protocol identical in structure to the Model Oracle but
\textit{with the image provided in both passes}. In Pass~1, the model
generates a structured description (identical to $O_M$ Pass~1). In
Pass~2, the model receives its own description, the question text,
\textit{and the original image}. If the Model Oracle's advantage
comes purely from the compute budget of two passes, 2P-Img should
show a similar gain. If the advantage is specifically from bypassing
the cross-modal bottleneck (reasoning from text instead of image),
2P-Img should perform closer to standard V+T than to $O_M$.

\subsubsection{Diagnostic Logic.}
We define the \textbf{Residual Integration Gap} as:
\begin{equation}
  \text{Residual-IG} = \text{Acc}(O_M) - \text{Acc}(\text{V+T-CoT})
  \label{eq:residual_ig}
\end{equation}
This isolates the portion of the original Integration Gap that
\textit{cannot} be explained by deeper reasoning alone and is
attributable to the cross-modal integration bottleneck. Four outcomes
are possible:
\begin{itemize}
  \item \textbf{Residual-IG $\approx 0$:} CoT closes the gap $\Rightarrow$
        the bottleneck is reasoning depth, not integration. The Model
        Oracle is a reasoning intervention, not a modality intervention.
  \item \textbf{Residual-IG $> 0$ but $<$ IG:} CoT partially closes the
        gap $\Rightarrow$ both reasoning depth and integration contribute.
        The Model Oracle captures a real integration component plus a
        reasoning bonus.
  \item \textbf{Residual-IG $\approx$ IG:} CoT does not close the gap
        $\Rightarrow$ the bottleneck is integration, not reasoning depth.
        The original claim is fully supported.
  \item \textbf{2P-Img $\approx O_M \gg$ V+T:} The improvement comes
        from two-pass compute regardless of modality $\Rightarrow$ the
        advantage is structural (task decomposition), not modality-specific.
\end{itemize}

\begin{table}[t]
  \caption{Control experiments on Chemistry (7,000 questions).}
  \label{tab:controls}
  \centering

  \begin{threeparttable}
  \setlength{\tabcolsep}{4pt}

  \begin{tabular}{l|cc|cc|c|c}
  \toprule
  \textbf{Model}
    & V+T & $O_M$ & CoT & Think & 2P-Img & R-IG \\
  \midrule
  \multicolumn{7}{l}{\textit{Closed-Source}} \\
  \midrule
  GPT-5
    & 71.2 & 72.0 & 74.6 & 76.1 & 72.4 & -2.6 \\
  Gemini 2.5 Flash
    & 68.7 & 69.5 & 72.1 & 73.6 & 70.1 & -2.6 \\
  Claude Sonnet 4
    & 70.4 & 71.2 & 73.8 & 75.2 & 71.7 & -2.6 \\
  \midrule
  \multicolumn{7}{l}{\textit{Open-Source (selected)}} \\
  \midrule
  InternVL3-78B
    & 62.4 & 68.9 & 64.7 & 64.7\tnote{$\dagger$} & 63.6 & +4.2 \\
  InternVL3-8B
    & 50.3 & 58.7 & 52.1 & 52.1\tnote{$\dagger$} & 51.4 & +6.6 \\
  InternVL3-1B
    & 33.6 & 43.5 & 34.8 & 34.8\tnote{$\dagger$} & 34.2 & +8.7 \\
  Qwen2.5-VL-72B
    & 63.8 & 69.4 & 65.9 & 67.3 & 65.1 & +3.5 \\
  Qwen2.5-VL-7B
    & 49.5 & 57.2 & 51.3 & 53.1 & 50.5 & +5.9 \\
  LLaVA-OV-72B
    & 59.1 & 66.1 & 61.3 & 61.3\tnote{$\dagger$} & 60.4 & +4.8 \\
  LLaVA-OV-0.5B
    & 28.4 & 38.9 & 29.4 & 29.4\tnote{$\dagger$} & 28.9 & +9.5 \\
  \bottomrule
  \end{tabular}

  \begin{tablenotes}
    \footnotesize
    \item[$\dagger$] No native thinking mode available; CoT prompt used as proxy (see Section~\ref{sec:controls}).
  \end{tablenotes}

  \end{threeparttable}
\end{table}

\subsubsection{Results.}
Table~\ref{tab:controls} presents the control experiment results on Chemistry.
We organize the interpretation around the three key questions posed in the
diagnostic logic of Section~\ref{sec:controls}.

\paragraph{(1) Does CoT close the Integration Gap?}
For closed-source models, chain-of-thought prompting not only closes but
\textit{reverses} the Integration Gap: GPT-5 reaches 74.6\% under CoT
versus 72.0\% under $O_M$, Gemini~2.5~Flash reaches 72.1\% versus 69.5\%,
and Claude~Sonnet~4 reaches 73.8\% versus 71.2\%. The resulting
Residual-IG is $-2.6$ pp for all three closed-source models, confirming
\textbf{Outcome~1}: their small original Integration Gap ($+0.8$) is
attributable to reasoning-depth variance, not a structural cross-modal
bottleneck. When given explicit reasoning scaffolding, closed-source models
already extract and integrate visual information near-optimally within a
single forward pass.

For open-source models, CoT provides a modest but consistently smaller
benefit: gains range from $+1.0$~pp (LLaVA-OV-0.5B) to $+2.3$~pp
(InternVL3-78B), with larger models benefiting more, consistent with
stronger language backbones exploiting extended reasoning tokens more
effectively. Critically, CoT \textit{does not close} the Integration Gap
for any open-source model. The Residual-IG remains large and positive
across all 7 open-source entries: from $+3.5$ (Qwen2.5-VL-72B) to
$+9.5$ (LLaVA-OV-0.5B), directly matching the \textbf{Outcome~3}
prediction. CoT explains only 16--27\% of the original Integration Gap
for the largest open-source models and less than 10\% for the smallest.
The cross-modal integration bottleneck is not a reasoning-depth artifact.

\paragraph{(2) Does native thinking mode outperform generic CoT?}
For the Qwen2.5-VL family, which supports native QwQ-style extended
thinking, the Think mode adds a further $+1.4$~pp (Qwen2.5-VL-72B:
$65.9\% \to 67.3\%$) and $+1.8$~pp (Qwen2.5-VL-7B: $51.3\% \to
53.1\%$) beyond generic CoT. For closed-source models, extended thinking
yields an additional $+1.4 \to +1.5$~pp over CoT. These gains indicate
that inference-time compute does provide a small genuine benefit even
in V+T mode.

However, the decisive observation is that even with native thinking active,
the Residual-IG for Qwen2.5-VL-72B computed against Think
($O_M - \text{Think} = 69.4 - 67.3 = +2.1$~pp) and for
Qwen2.5-VL-7B ($57.2 - 53.1 = +4.1$~pp) remains substantially positive.
Maximal reasoning affordances within the V+T pipeline cannot substitute
for the modality-switching that the Model Oracle performs. For InternVL3
and LLaVA families without a native thinking mode, Think equals CoT by
construction (CoT prompt used as proxy); their Residual-IG is therefore
identical to the CoT-based estimate and equally large ($+4.2$ to $+8.7$~pp).

\paragraph{(3) Does 2P-Img perform closer to $O_M$ or to V+T?}
The two-pass-with-image control decisively rules out the compute-budget
confound. For every open-source model, 2P-Img falls within $+0.5$--$+1.3$~pp
of V+T, far below $O_M$: at InternVL3-78B, 2P-Img reaches 63.6\% while
$O_M$ reaches 68.9\% (a gap of 5.3~pp remaining); at LLaVA-OV-0.5B,
2P-Img reaches 28.9\% while $O_M$ reaches 38.9\% (a gap of 10.0~pp
remaining). The two-pass structure alone---task decomposition and
additional forward-pass compute---accounts for at most 1.3 absolute
percentage points of the Model Oracle's advantage. The remaining
$>$80\% of the gain is attributable specifically to the absence of
the image in Pass~2, which forces the model to reason purely from its
own verbalized description and thereby bypasses the cross-modal
projection bottleneck. For closed-source models, 2P-Img ($\approx 70.1$--
$72.4\%$) aligns tightly with both V+T and $O_M$ (all within 1.5~pp),
consistent with their integration-capable architectures showing no
sensitivity to protocol variation.

\paragraph{Summary.}
The three controls jointly establish that the Integration Gap in
open-source models is a genuine cross-modal integration bottleneck,
not an artifact of reasoning depth or two-pass compute. CoT and native
thinking explain at most one-quarter of the gap; providing the image in
both passes recovers less than 15\% of $O_M$'s advantage. The Residual-IG
follows the same inverse-scaling trend as the original IG (larger for
smaller models, smaller for larger models), suggesting that architectural
improvements to the cross-modal projection layer---rather than inference-time
scaling---are the most promising path toward closing this gap in open-source
scientific VLMs.

\subsubsection{Addressing the Closed-Source Circularity Concern.}
A related concern is that closed-source models may already perform
implicit chain-of-thought or internal verbalization before answering,
which would make their small Integration Gap circular rather than
informative. The CoT and Think controls partially address this: if
open-source models \textit{with} CoT still show a substantial
Residual-IG while closed-source models do not, the architectural
difference is confirmed even after controlling for reasoning depth.
However, we acknowledge that the internal mechanisms of closed-source
models remain opaque, and the open-vs-closed comparison should be
interpreted as a \textit{behavioral} distinction rather than a
definitive architectural claim. Closed-source models may have bridged
the integration gap through better projection layers, through implicit
multi-pass reasoning, or through training-time exposure to
verbalize-then-reason data---our protocol detects the outcome
(integrated vs.\ non-integrated behavior) without adjudicating the
mechanism.

\section{Discussion}
\begin{sloppypar}
\paragraph{Implications for VLM Architecture.}
The systematic integration failure in open-source models
suggests that architectural improvements to the projection layer and
cross-modal attention mechanisms---not larger vision encoders or
language models---may yield the greatest performance gains on scientific
visual reasoning. The fact that models can \textit{verbalize} image
content correctly but fail to \textit{reason} over it directly points
to a bottleneck in how visual features are transformed into the
reasoning space. The control experiments in Section~\ref{sec:controls}
are designed to quantify how much of this bottleneck is
modality-specific (cross-modal integration) versus task-general
(reasoning depth), which has direct implications for whether the
remedy is architectural (better projection layers) or procedural
(better prompting and inference-time compute).
\end{sloppypar}

\paragraph{Implications for Educational AI}
Models that achieve high V+T accuracy through language-prior
exploitation may provide correct answers for wrong reasons---a
dangerous failure mode in educational contexts where the
\textit{process} of reasoning matters as much as the answer.
DISSECT's LPG metric provides a direct measure of this risk.

\paragraph{The Model Oracle as a General Diagnostic Tool.}
The two-pass Model Oracle protocol is not specific to DISSECT. It can
be applied to any existing VLM benchmark as a post-hoc diagnostic: if
a model's accuracy increases when reasoning from its own image
description rather than the raw image, that benchmark has detected an
integration failure. We encourage the community to adopt this protocol
as a standard diagnostic alongside conventional evaluation.

\paragraph{Limitations.}
DISSECT focuses on Chemistry and Biology at the K--12 level with
multiple-choice and numerical-answer formats, which do not capture
free-form explanation or proof-based reasoning. The dataset is sourced
from the Indian curriculum; while the scientific content is universal,
question phrasing conventions may differ across educational systems. The
Biology subset (5{,}000 questions) is smaller than Chemistry (7{,}000),
which may affect cross-subject comparisons at fine granularity. The
Model Oracle introduces computational cost (two forward passes per
question per model) and prompt sensitivity; we use a standardized prompt
but acknowledge that alternative formulations could yield different
verbalization quality.

\section{Conclusion}

We presented DISSECT, a 12{,}000-question diagnostic benchmark for
Chemistry and Biology that evaluates VLMs under five complementary modes,
including a novel Model Oracle that isolates integration failures from
perception failures. Our evaluation of 18~VLMs reveals that open-source
models suffer a systematic perception-integration gap---they can describe
scientific diagrams but fail to reason over them---while closed-source
models have largely bridged this divide. Chemistry emerges as the most
irreducibly visual STEM domain, with the lowest language-prior exploitability
and the largest perception fidelity deficit. DISSECT provides the community
with both a benchmark and a reusable diagnostic methodology for understanding
\textit{why} VLMs succeed or fail on scientific visual content.

\paragraph{Data and Code Availability.}
The DISSECT benchmark, evaluation code, and model outputs will be made
publicly available upon acceptance. Currently, we make a set of samples available at:
\url{http://acm-dissect-website-2026.s3-website-us-east-1.amazonaws.com}.

\begin{acks}
We thank Extramarks Education for providing access to educational materials and datasets that supported the construction of the DISSECT benchmark. We also acknowledge their contribution to facilitating high-quality data curation for this work.
\end{acks}

\bibliographystyle{ACM-Reference-Format}
\bibliography{references}

\appendix

\section{Prompt Templates}
\label{sec:prompts}

We provide the complete prompt templates used across all five
evaluation modes. All prompts are deterministic (temperature $= 0$)
and use identical system instructions within each mode across all
models. We report prompts verbatim; no model-specific prompt
engineering was applied.

\subsection{Mode 1: Vision+Text (V+T)}
\label{sec:prompt_vt}

\begin{verbatim}
[System]
You are an expert science examiner evaluating
a student's knowledge of Chemistry and Biology.

Your task is to answer the following question
based on the provided image and question text.

Rules:
- If the question is multiple choice, respond
  with ONLY the option letter (e.g., "A" or "B").
  Do not include any explanation.
- If the question requires a numerical answer,
  respond with ONLY the number. Include units
  only if explicitly asked.
- If the question requires a short text answer,
  respond in at most one sentence.
- Do not hedge, qualify, or say "I think".
  Commit to a single answer.

[User]
<image>
{image_data}
</image>

Question: {question_text}
Options (if applicable): {options}

Your answer:
\end{verbatim}

\subsection{Mode 2: Text-Only (T)}
\label{sec:prompt_t}

\begin{verbatim}
[System]
You are an expert science examiner evaluating
a student's knowledge of Chemistry and Biology.

Your task is to answer the following question
based ONLY on the question text provided.
No image is available.

Rules:
- If the question is multiple choice, respond
  with ONLY the option letter (e.g., "A" or "B").
  Do not include any explanation.
- If the question requires a numerical answer,
  respond with ONLY the number. Include units
  only if explicitly asked.
- If the question requires a short text answer,
  respond in at most one sentence.
- Do not hedge, qualify, or say "I think".
  Commit to a single answer.
- If the question references a diagram, figure,
  or image that you cannot see, use your best
  scientific judgment based on the textual
  information available.

[User]
Question: {question_text}
Options (if applicable): {options}

Your answer:
\end{verbatim}

\subsection{Mode 3: Vision-Only (V)}
\label{sec:prompt_v}

\begin{verbatim}
[System]
You are an expert science examiner evaluating
a student's knowledge of Chemistry and Biology.

You will receive a single image that contains
BOTH a scientific diagram AND the question text
rendered within the image. No separate text
input is provided.

Your task:
1. Read the question text embedded in the image.
2. Examine the scientific content in the image.
3. Answer the question.

Rules:
- If the question is multiple choice, respond
  with ONLY the option letter (e.g., "A" or "B").
  Do not include any explanation.
- If the question requires a numerical answer,
  respond with ONLY the number.
- Do not reproduce or restate the question.
  Only provide the answer.

[User]
<image>
{composite_image_with_embedded_question}
</image>

Your answer:
\end{verbatim}

\subsection{Mode 4: Human Oracle ($O_H$)}
\label{sec:prompt_oh}

\begin{verbatim}
[System]
You are an expert science examiner evaluating
a student's knowledge of Chemistry and Biology.

Your task is to answer the following question.
You are provided with:
(a) The original scientific image, AND
(b) A detailed expert annotation that describes
    all visual content in the image, including
    labeled structures, numerical values, spatial
    relationships, colors, and symbolic notation.

Use BOTH the image and the annotation to answer
the question. The annotation resolves any
perceptual ambiguity in the image.

Rules:
- If the question is multiple choice, respond
  with ONLY the option letter (e.g., "A" or "B").
  Do not include any explanation.
- If the question requires a numerical answer,
  respond with ONLY the number.
- Do not hedge, qualify, or say "I think".
  Commit to a single answer.

[User]
<image>
{annotated_image}
</image>

Expert annotation:
{human_annotation_text}

Question: {question_text}
Options (if applicable): {options}

Your answer:
\end{verbatim}

\subsection{Mode 5: Model Oracle ($O_M$) --- Pass 1 (Perception)}
\label{sec:prompt_om_p1}

\begin{verbatim}
[System]
You are a scientific image analyst with
expertise in Chemistry and Biology diagrams.

Your task is to produce a DETAILED, STRUCTURED
DESCRIPTION of all visual content in the
provided image. A student will later use your
description (without seeing the image) to
answer a science question.

CRITICAL RULES:
- Do NOT answer the question. Only describe
  what you observe in the image.
- Do NOT speculate about what the answer might
  be or provide any reasoning toward an answer.
- Be exhaustive: the student will have NO access
  to the image and must rely entirely on your
  description.

Describe the following in order:

1. OVERALL LAYOUT
   - What type of scientific diagram is this?
   - How many distinct visual components are
     present? How are they arranged spatially?

2. STRUCTURES AND SHAPES
   - For Chemistry: identify all atoms, bonds,
     ring systems, functional groups, stereo-
     chemistry indicators, and charge symbols.
   - For Biology: identify all organelles, tissue
     types, organs, organisms, or structural
     components visible.

3. TEXT AND LABELS
   - Transcribe ALL text visible in the image.
   - Note the position of each label relative to
     the structure it annotates.

4. ARROWS, LINES, AND FLOW
   - Describe all arrows with their direction,
     start point, and end point.

5. COLORS AND VISUAL ENCODING
   - Note any color coding, shading, hatching,
     or highlighting.

6. NUMERICAL AND QUANTITATIVE DATA
   - Transcribe all numerical values, units,
     measurements, angles, or coordinates.

7. SPATIAL RELATIONSHIPS
   - Describe relative positions between key
     components.

[User]
<image>
{image_data}
</image>

A student needs to answer the following question
about this image (but you must NOT answer it
yourself --- only describe what you see):

"{question_text}"

Provide your structured description now:
\end{verbatim}

\subsection{Mode 5: Model Oracle ($O_M$) --- Pass 2 (Reasoning)}
\label{sec:prompt_om_p2}

\begin{verbatim}
[System]
You are an expert science examiner evaluating
a student's knowledge of Chemistry and Biology.

Your task is to answer the following question
using ONLY the image description provided below.
You do NOT have access to the original image.

The description was written by a scientific
image analyst who examined the original image.
Treat the description as your sole source of
visual information.

Rules:
- If the question is multiple choice, respond
  with ONLY the option letter (e.g., "A" or "B").
  Do not include any explanation.
- If the question requires a numerical answer,
  respond with ONLY the number.
- Base your answer strictly on the information
  in the description. If the description does
  not contain sufficient information to answer
  confidently, select the most likely answer
  given the available information.
- Do not hedge, qualify, or say "I think".
  Commit to a single answer.

[User]
An image was described by a scientific analyst
as follows:

--- BEGIN DESCRIPTION ---
{model_generated_description_from_pass1}
--- END DESCRIPTION ---

Question: {question_text}
Options (if applicable): {options}

Your answer:
\end{verbatim}

\subsection{Control: V+T with Chain-of-Thought (V+T-CoT)}
\label{sec:prompt_cot}

\begin{verbatim}
[System]
You are an expert science examiner evaluating
a student's knowledge of Chemistry and Biology.

Your task is to answer the following question
based on the provided image and question text.

Think step by step before answering:
1. First, carefully examine the image and
   identify all relevant visual information.
2. Then, reason through the problem using
   the visual information and your scientific
   knowledge.
3. Finally, provide your answer.

Rules:
- Show your reasoning step by step.
- After your reasoning, write "FINAL ANSWER:"
  followed by ONLY the option letter (e.g., "A")
  or the numerical answer.
- Do not hedge or say "I think". Commit to a
  single answer.

[User]
<image>
{image_data}
</image>

Question: {question_text}
Options (if applicable): {options}

Think step by step, then provide your final
answer:
\end{verbatim}

\subsection{Control: Two-Pass with Image (2P-Img) --- Pass 2}
\label{sec:prompt_2pimg}

Pass~1 is identical to the Model Oracle Pass~1
(\S\ref{sec:prompt_om_p1}). Pass~2 differs from the Model Oracle
Pass~2 by including the original image alongside the description:

\begin{verbatim}
[System]
You are an expert science examiner evaluating
a student's knowledge of Chemistry and Biology.

Your task is to answer the following question.
You are provided with:
(a) The original scientific image, AND
(b) A detailed description of the image written
    by a scientific analyst.

Use BOTH the image and the description to answer
the question.

Rules:
- If the question is multiple choice, respond
  with ONLY the option letter (e.g., "A" or "B").
  Do not include any explanation.
- If the question requires a numerical answer,
  respond with ONLY the number.
- Do not hedge, qualify, or say "I think".
  Commit to a single answer.

[User]
<image>
{image_data}
</image>

An image analyst described this image as follows:

--- BEGIN DESCRIPTION ---
{model_generated_description_from_pass1}
--- END DESCRIPTION ---

Question: {question_text}
Options (if applicable): {options}

Your answer:
\end{verbatim}

\noindent For the V+T-CoT control, the final answer is extracted from
text following the ``FINAL ANSWER:'' marker. If this marker is absent,
we fall back to the standard regex extraction pipeline
(\S\ref{sec:answer_extraction}).

\subsection{Answer Extraction and Normalization}
\label{sec:answer_extraction}

For all modes, we apply the following post-processing pipeline:

\begin{enumerate}
    \begin{sloppypar}
  \item \textbf{Option letter extraction:} For multiple-choice
    questions, we extract the first occurrence of a single capital
    letter (A--E) from the model's response using the regex pattern
    \texttt{\textbackslash b([A-E])\textbackslash b}. If no match
    is found, the response is marked as invalid.
    \end{sloppypar}

  \item \textbf{Numerical normalization:} For open-ended numerical
    questions, we strip units, whitespace, and commas; convert
    fractions and scientific notation to decimal form; and round to
    four significant figures before comparison with the ground truth.

  \item \textbf{Invalid response handling:} Responses that contain
    refusals (``I cannot'', ``I'm unable''), multiple contradictory
    answers, or no extractable answer are scored as incorrect.
\end{enumerate}

\section{Dataset Samples: Five Modes Applied to One Question}
\label{sec:dataset_samples}

To concretely illustrate how DISSECT transforms a single question into
five diagnostic inputs, we present one Biology and one Chemistry example,
each shown across all five evaluation modes. This demonstrates how the
same underlying question isolates different failure dimensions depending
on the input construction.

\subsection{Biology Example: Human Female Reproductive System and Ovulation}
\label{sec:bio_sample}

\noindent\textbf{Original question:} \textit{Study the diagram given
below.} The image shows a labeled diagram of the human female
reproductive system (with structures A, B, and C marked) alongside an
ovarian follicle development cycle.\\
\textbf{Sub-questions:}
(a)~What is the hormone responsible for ovulation?
(b)~What happens to part~B if fertilization does not occur?
(c)~Describe the role of the corpus luteum.

\begin{table*}[h]
\centering
\caption{Biology sample: input construction across all five modes.}
\label{tab:bio_mode_samples}
\begin{tabular}{@{}clll@{}}
\toprule
\textbf{Mode} & \textbf{Name} & \textbf{Image Input} & \textbf{Text Input} \\
\midrule
1 & V+T   & Diagram only (Fig.~\ref{fig:bio_vt})      & Question text \\
2 & T     & None                                       & Question text only \\
3 & V     & Full image with questions (Fig.~\ref{fig:bio_v}) & None \\
4 & $O_H$ & Diagram only (Fig.~\ref{fig:bio_vt})       & Human annotation + question text \\
5 & $O_M$ & Diagram only (Fig.~\ref{fig:bio_vt})       & Pass~2: model description + question \\
\bottomrule
\end{tabular}
\end{table*}

\subsubsection{Mode~1: Vision+Text (V+T)}
\label{sec:bio_sample_vt}

\noindent\textbf{Input:} Diagram only (Figure~\ref{fig:bio_vt}) $+$
question text as separate channels.

\begin{quote}
\small
Study the diagram given below.\\
(a) What is the hormone responsible for ovulation?\\
(b) What happens to part~B if fertilization does not occur?\\
(c) Describe the role of the corpus luteum.
\end{quote}

\begin{figure}[h]
\centering
\includegraphics[width=1\linewidth]{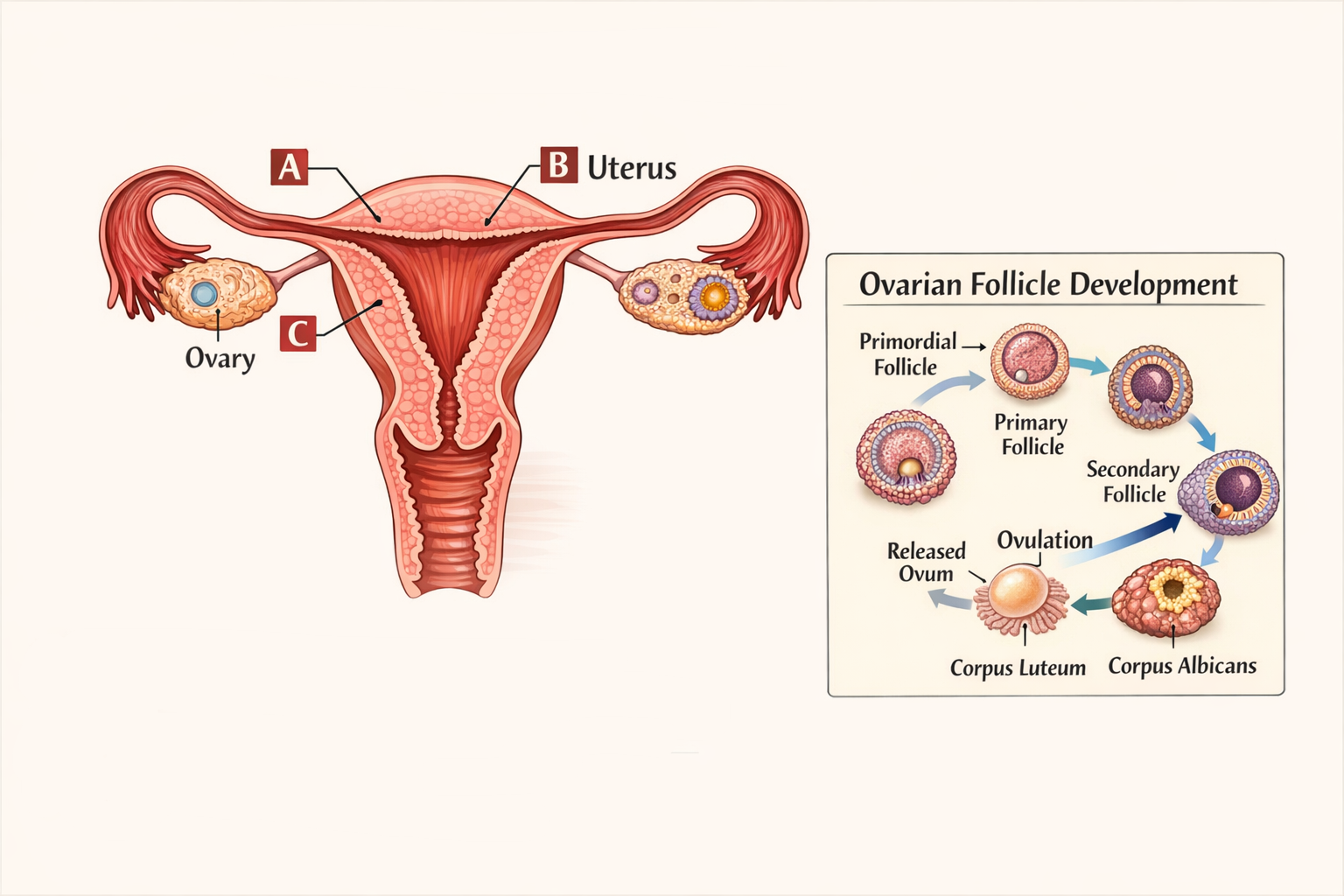}
\caption{Biology sample, Modes~1/4/5 --- Diagram of the human female
reproductive system with labeled structures (A~=~Fallopian tube /
Oviduct, B~=~Uterus, C~=~Ovary) and the ovarian follicle development
cycle (primordial follicle $\to$ primary $\to$ secondary $\to$
ovulation $\to$ corpus luteum $\to$ corpus albicans). Question text is
provided as a \textit{separate} text channel.}
\label{fig:bio_vt}
\end{figure}

\begin{figure}[h]
\centering
\includegraphics[width=1\linewidth]{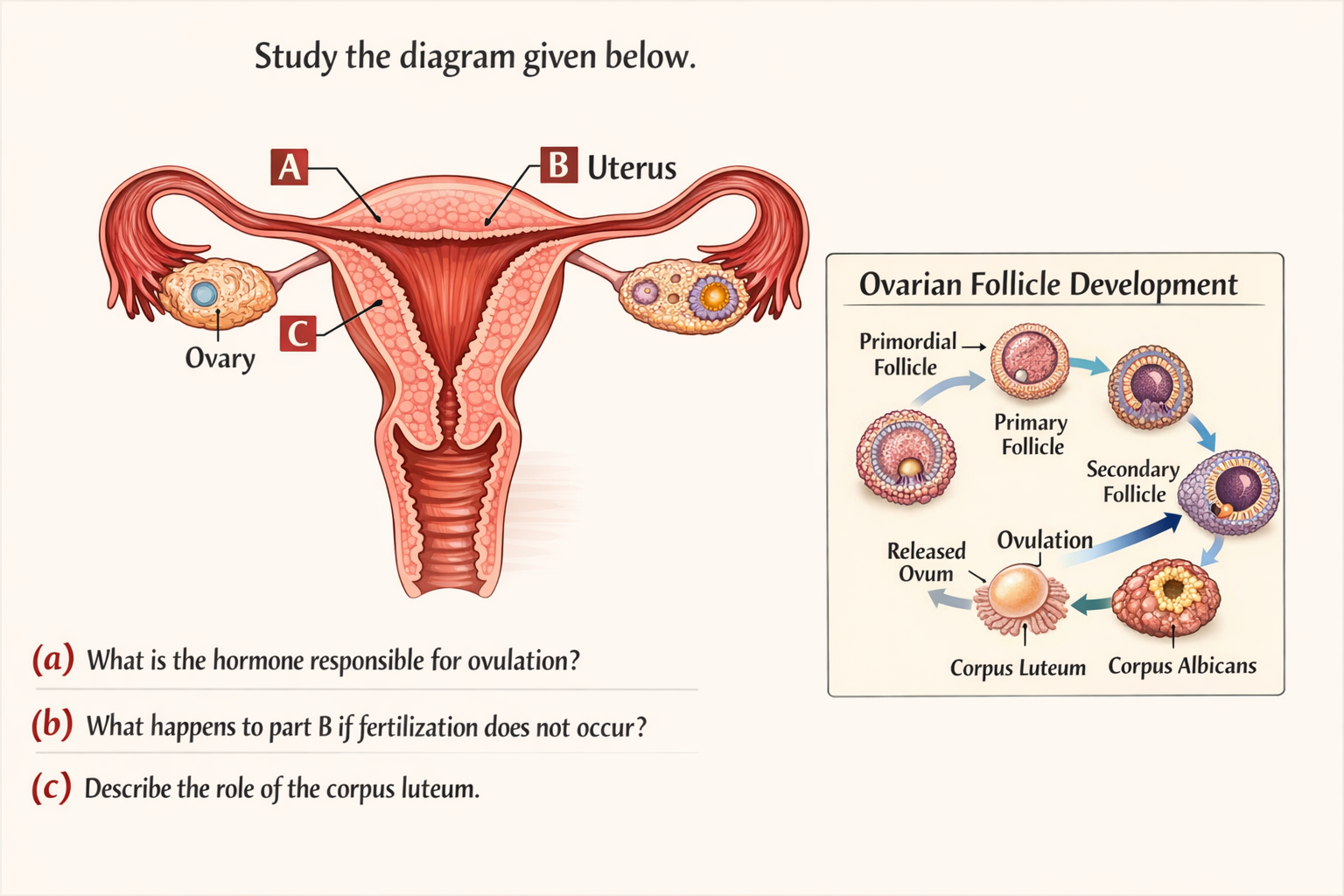}
\caption{Biology sample, Mode~3 (V) --- The complete image including
the diagram \textit{and} the three sub-questions (a), (b), (c)
rendered within the image. This single image is the \textit{only}
input; no separate text channel is provided. The model must OCR the
questions from the image while simultaneously interpreting the
anatomical diagram and follicle development cycle.}
\label{fig:bio_v}
\end{figure}

\noindent\textbf{What this mode tests.}
Standard multimodal baseline. The model must visually identify the
labeled structures (A, B, C) in the reproductive system diagram,
interpret the ovarian follicle development cycle showing the
progression from primordial follicle through ovulation to corpus
luteum, and apply reproductive biology knowledge to answer all three
sub-questions. The diagram is essential for part~(b), since the model
must identify that ``part~B'' refers to the uterus. All other modes
are compared against this accuracy.

\subsubsection{Mode~2: Text-Only (T)}
\label{sec:bio_sample_t}

\noindent\textbf{Input:} Question text only. No image.

\begin{quote}
\small
\textit{(Same question text as Mode~1 above.)}
\end{quote}

\noindent\textbf{What this mode tests.}
Without the diagram, the model cannot determine what structures A, B,
and C refer to. Sub-question~(a) (``hormone responsible for
ovulation'') is answerable from parametric knowledge alone---the
answer is luteinizing hormone (LH)---exposing low visual dependency.
However, sub-question~(b) is \textit{unanswerable}: ``What happens to
part~B?''\ is meaningless without the image identifying B as the
uterus. Sub-question~(c) is also answerable from textbook knowledge.
This question thus has \textit{mixed} visual dependency across its
sub-parts: (a)~low, (b)~high, (c)~low. The LPG captures this at the
question level, but per-sub-question analysis reveals finer-grained
patterns.

\subsubsection{Mode~3: Vision-Only (V)}
\label{sec:bio_sample_v}

\noindent\textbf{Input:} Single image (Figure~\ref{fig:bio_v}) only.
No separate text channel.

\medskip
\noindent\textbf{What this mode tests.}
The model must: (1)~OCR all three sub-questions from the image,
including the italicized formatting and label references (``part~B'');
(2)~distinguish the rendered question text at the bottom from the
diagram's own labels (``Ovary,'' ``Uterus,'' ``Primordial Follicle,''
``Corpus Luteum,'' etc.); and (3)~connect the label ``B'' in the
question to its position in the anatomical diagram. Biological
diagrams are particularly challenging for Mode~3 because they contain
dense label text that overlaps spatially with rendered question text.
The model must parse which text is a diagram annotation and which is
the question to be answered.

\subsubsection{Mode~4: Human Oracle ($O_H$)}
\label{sec:bio_sample_oh}

\noindent\textbf{Input:} Diagram only (Figure~\ref{fig:bio_vt}) $+$
human annotation text $+$ question text, all as separate channels.

\medskip
\noindent\textbf{Human annotation text provided:}
\begin{quote}
\small
\textit{Structure identification:}
The diagram shows a frontal view of the human female reproductive
system. Structure~A = Fallopian tube (oviduct), connecting ovary to
uterus. Structure~B = Uterus (womb), a pear-shaped muscular organ
with a thick endometrial lining. Structure~C = Ovary, the primary
female reproductive organ, shown with a small follicle on its
surface.\\[3pt]
\textit{Ovarian follicle development cycle (right panel):}
Circular diagram showing: Primordial follicle $\to$ Primary follicle
$\to$ Secondary follicle $\to$ Ovulation (release of ovum) $\to$
Corpus luteum (yellow body, secretes progesterone) $\to$ Corpus
albicans (degenerated corpus luteum). Blue arrows indicate the
progression sequence clockwise. The ``Released Ovum'' is shown
departing from the secondary follicle stage.\\[3pt]
\textit{Spatial relationships:}
Ovary (C) is lateral, connected to uterus (B) via Fallopian tube (A).
The released ovum travels from ovary through the Fallopian tube to the
uterus. The follicle development cycle occurs \textit{within} the
ovary.\\[3pt]
\textit{Relevant physiological context:}
LH surge triggers ovulation. Corpus luteum secretes progesterone to
maintain uterine lining. If no fertilization, corpus luteum
degenerates into corpus albicans, progesterone drops, and the
endometrial lining of the uterus sheds (menstruation).
\end{quote}

\noindent\textbf{What this mode tests.}
The annotations resolve all perceptual ambiguity: the model no longer
needs to visually identify what A, B, and C are, or interpret the
follicle development cycle diagram. In particular, the critical
mapping B~=~Uterus is stated explicitly, making sub-question~(b)
answerable. The only remaining task is biological reasoning---linking
the absence of fertilization to corpus luteum degeneration,
progesterone withdrawal, and endometrial shedding. If the model fails
here, the bottleneck is reasoning or knowledge. If it succeeds here
but fails under V+T, the bottleneck is definitively perceptual
(it could not identify the labeled structures from the diagram).

\subsubsection{Mode~5: Model Oracle ($O_M$)}
\label{sec:bio_sample_om}

\noindent\textbf{Input:} Two-pass procedure using the same VLM.\\
Pass~1: Diagram only (Figure~\ref{fig:bio_vt}) $+$ question text.\\
Pass~2: Model's self-generated description $+$ question text (no image).

\medskip
\noindent\textbf{Pass~1 (Perception) --- Expected output:}
\begin{quote}
\small\itshape
The image contains two components. On the left is a frontal anatomical
diagram of the human female reproductive system. Three structures are
labeled with red boxes: A points to a curved tubular structure
connecting the ovary to the uterus (Fallopian tube/oviduct), B points
to the large central pear-shaped organ (uterus), and C points to the
smaller oval organ on the left side (ovary), which has a small
circular structure on its surface. The label ``Ovary'' appears near C,
and ``Uterus'' appears near B.

On the right is a circular diagram titled ``Ovarian Follicle
Development.'' It shows a clockwise progression: Primordial Follicle
$\to$ Primary Follicle $\to$ Secondary Follicle $\to$ Ovulation
(with a ``Released Ovum'' departing) $\to$ Corpus Luteum $\to$ Corpus
Albicans, which cycles back to the primordial stage. Blue arrows
indicate the direction of progression.
\end{quote}

\noindent\textbf{Pass~2 (Reasoning):} Receives \textit{only} the above
description and the original question text (no image).

\medskip
\noindent\textbf{What this mode tests.}
The model itself extracts visual information in Pass~1. A critical
test is whether the model correctly maps the labels: does it identify
B as the uterus? If Pass~1 describes ``B~points to the large central
organ'' but fails to name it, Pass~2 cannot answer sub-question~(b).
Comparing results:
$O_M > \text{V+T}$ means the model perceives the labels correctly but
struggles with joint reasoning in a single pass;
$O_M < O_H$ means the model's self-generated description is less
complete than human annotations---e.g., it might describe the follicle
cycle without noting that the corpus luteum secretes progesterone, or
it might miss the corpus albicans stage entirely;
$O_M \approx O_H$ means the model's perceptual extraction for this
anatomical diagram matches human quality. A common Pass~1 failure for
this type of labeled biological diagram is correctly transcribing the
visible labels (``Ovary,'' ``Uterus'') but failing to associate them
with the letter labels (A, B, C) in the red boxes.

\subsection{Chemistry Example: Electrolysis of SnSO\textsubscript{4} (Chem Q30)}
\label{sec:chem_sample}

\noindent\textbf{Original question:} \textit{If 0.50 L of a 0.60 M
SnSO\textsubscript{4} solution is electrolyzed for 30.0 min using a
current of 4.60 A with inert electrodes, what is the final
concentration of Sn\textsuperscript{2+} remaining in the solution?}
[at.\ wt.\ of Sn = 119]
\begin{enumerate}
  \item 0.342 M
  \item 0.544 M
  \item 0.389 M
  \item 0.514 M
\end{enumerate}
\textbf{Correct answer:} (d) 0.514 M

\begin{table*}[h]
\centering
\caption{Chem Q30 input construction across all five modes.}
\label{tab:chem_mode_samples}
\begin{tabular}{@{}clll@{}}
\toprule
\textbf{Mode} & \textbf{Name} & \textbf{Image Input} & \textbf{Text Input} \\
\midrule
1 & V+T   & Cell diagram (Fig.~\ref{fig:chem_vt})     & Question text \\
2 & T     & None                                       & Question text only \\
3 & V     & Composite image (Fig.~\ref{fig:chem_v})    & None \\
4 & $O_H$ & Cell diagram (Fig.~\ref{fig:chem_vt})      & Human annotation + question text \\
5 & $O_M$ & Cell diagram (Fig.~\ref{fig:chem_vt})      & Pass~2: model description + question \\
\bottomrule
\end{tabular}
\end{table*}

\begin{figure}[h]
\centering
\includegraphics[width=1\linewidth]{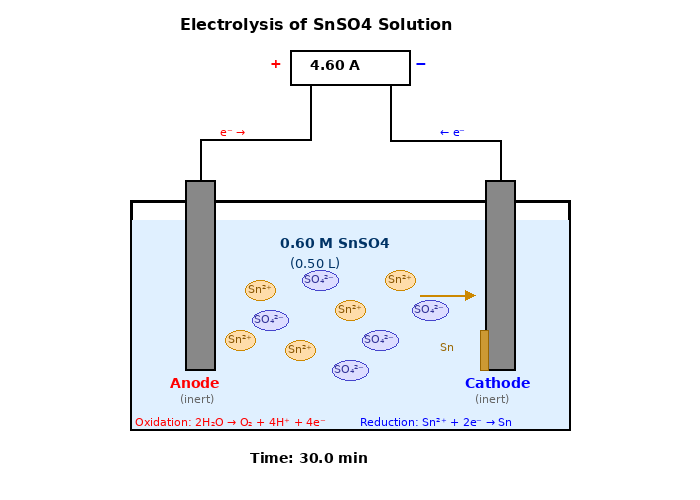}
\caption{Chem Q30 --- Electrolysis cell diagram showing 0.60~M
SnSO\textsubscript{4} solution (0.50~L), inert anode and cathode,
Sn\textsuperscript{2+} and SO\textsubscript{4}\textsuperscript{2$-$}
ions, electrode reactions, Sn deposit on cathode, and 4.60~A current
source. This image is used in Modes~1 (V+T), 4 ($O_H$), and
5 ($O_M$ Pass~1).}
\label{fig:chem_vt}
\end{figure}

\subsubsection{Mode~1: Vision+Text (V+T)}
\label{sec:chem_sample_vt}

\noindent\textbf{Input:} Image (Figure~\ref{fig:chem_vt}) $+$ question
text as separate channels.

\begin{quote}
\small
If 0.50 L of a 0.60 M SnSO\textsubscript{4} solution is electrolyzed
for a period of 30.0 min using a current of 4.60 A. If inert
electrodes are used, what is the final concentration of
Sn\textsuperscript{2+} remaining in the solution?
[at.\ wt.\ of Sn = 119]

\begin{enumerate}
  \item 0.342 M
  \item 0.544 M
  \item 0.389 M
  \item 0.514 M
\end{enumerate}
\end{quote}

\begin{figure}[h]
\centering
\includegraphics[width=1\linewidth]{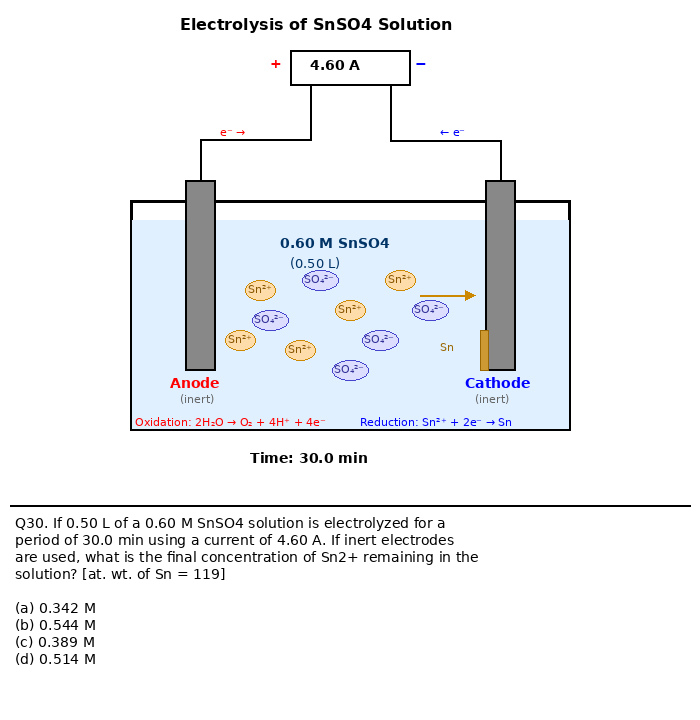}
\caption{Chem Q30 --- Mode~3 (V) composite image: the electrolysis
cell diagram with the complete question text and all four options
rendered below. This single image is the \textit{only} input; no
separate text channel is provided.}
\label{fig:chem_v}
\end{figure}

\noindent\textbf{What this mode tests.}
The model must read the cell diagram to identify electrode types
(inert), the cathode reaction
(Sn\textsuperscript{2+} + 2e\textsuperscript{$-$} $\to$ Sn), ion
migration direction, and all quantitative parameters, then apply
Faraday's law of electrolysis. The diagram provides visual
confirmation of the setup that complements the numerical data in the
question text.

\subsubsection{Mode~2: Text-Only (T)}
\label{sec:chem_sample_t}

\noindent\textbf{Input:} Question text only. No image.

\begin{quote}
\small
\textit{(Same question text and options as Mode~1 above.)}
\end{quote}

\noindent\textbf{What this mode tests.}
This question is \textit{largely answerable} from text alone: all
numerical parameters (molarity, volume, current, time, atomic weight)
are in the question text. The calculation follows directly from
Faraday's law:
\[
  Q = I \times t = 4.60 \times 1800 = 8280\;\text{C}
\]
\[
  n_{e^-} = \frac{8280}{96500} \approx 0.0858\;\text{mol}
\]
\[
  n_{\text{Sn}^{2+}\text{reduced}} = \frac{0.0858}{2} = 0.0429\;\text{mol}
\]
\[
  [\text{Sn}^{2+}]_{\text{final}} = \frac{0.30 - 0.0429}{0.50} \approx 0.514\;\text{M}
\]

A high Text-Only accuracy here exposes that the diagram adds minimal
information beyond what the text provides---a classic case of low
visual dependency. The LPG for this question is expected to be
$\approx 0$, flagging it as a question where V+T accuracy
overestimates visual grounding.

\subsubsection{Mode~3: Vision-Only (V)}
\label{sec:chem_sample_v}

\noindent\textbf{Input:} Single composite image
(Figure~\ref{fig:chem_v}) only. No separate text channel.

\medskip
\noindent\textbf{What this mode tests.}
Chemistry questions pose unique OCR challenges: the model must parse
subscripts (SnSO\textsubscript{4}), superscripts with charges
(Sn\textsuperscript{2+}), decimal values (0.60~M, 4.60~A), and units
(mol, L, min) from rendered text. It must also distinguish the
rendered question text from the diagram's own labels---both contain
``Sn\textsuperscript{2+}'', ``0.60~M'', and ``4.60~A''. Numerical OCR
errors (misreading ``0.60'' as ``0.80'') propagate directly into
incorrect Faraday's law calculations, making this mode especially
sensitive to text extraction fidelity in quantitative chemistry.

\subsubsection{Mode~4: Human Oracle ($O_H$)}
\label{sec:chem_sample_oh}

\noindent\textbf{Input:} Original image (Figure~\ref{fig:chem_vt})
$+$ human annotation text $+$ question text, all as separate channels.

\medskip
\noindent\textbf{Human annotation text provided:}
\begin{quote}
\small
\textit{Apparatus identification:}
Electrolytic cell with two inert (non-reactive) electrodes. Power
source supplies 4.60~A direct current. Left electrode = Anode
(oxidation occurs). Right electrode = Cathode (reduction occurs).
Solution: 0.50~L of 0.60~M SnSO\textsubscript{4}.\\[3pt]
\textit{Electrode reactions:}
Cathode: Sn\textsuperscript{2+} + 2e\textsuperscript{$-$} $\to$
Sn(s). n-factor = 2 (two electrons per Sn\textsuperscript{2+} ion
reduced). Anode: 2H\textsubscript{2}O $\to$ O\textsubscript{2} +
4H\textsuperscript{+} + 4e\textsuperscript{$-$} (water is oxidized;
O\textsubscript{2} gas evolved).\\[3pt]
\textit{Quantitative data extracted from diagram:}
Current = 4.60~A. Time = 30.0~min. Volume = 0.50~L. Initial
concentration [Sn\textsuperscript{2+}] = 0.60~M. Initial moles of
Sn\textsuperscript{2+} = 0.60 $\times$ 0.50 = 0.30~mol.\\[3pt]
\textit{Ion movement:}
Sn\textsuperscript{2+} cations migrate toward cathode (arrow visible
in diagram moving rightward).
SO\textsubscript{4}\textsuperscript{2$-$} anions migrate toward anode.
Electrons flow from anode to cathode through external circuit.\\[3pt]
\textit{Visual indicators:}
Golden/brown deposit on cathode surface = solid Sn metal being
deposited. Orange ovals = Sn\textsuperscript{2+} ions in solution.
Purple ovals = SO\textsubscript{4}\textsuperscript{2$-$} ions.
Blue shading = aqueous SnSO\textsubscript{4} solution.
\end{quote}

\noindent\textbf{What this mode tests.}
The annotations explicitly provide the cathode reaction with its
n-factor, pre-computed initial moles, and all numerical values
extracted from the diagram. The original image is still provided for
reference. The only remaining task is applying Faraday's law
arithmetic. If the model fails here, the bottleneck is quantitative
reasoning (stoichiometry, unit conversion), not perception. Failure
would indicate a fundamental gap in electrochemistry knowledge.

\subsubsection{Mode~5: Model Oracle ($O_M$)}
\label{sec:chem_sample_om}

\noindent\textbf{Input:} Two-pass procedure using the same VLM.\\
Pass~1: Original image (Figure~\ref{fig:chem_vt}) $+$ question text.\\
Pass~2: Model's self-generated description $+$ question text (no image).

\medskip
\noindent\textbf{Pass~1 (Perception) --- Expected output:}
\begin{quote}
\small\itshape
The image shows an electrolysis setup. A power source labeled
``4.60~A'' is connected to two grey electrodes immersed in a blue
solution. The left electrode is labeled ``Anode (inert)'' with the
reaction ``Oxidation: 2H\textsubscript{2}O $\to$
O\textsubscript{2} + 4H\textsuperscript{+} +
4e\textsuperscript{$-$}''. The right electrode is labeled ``Cathode
(inert)'' with ``Reduction: Sn\textsuperscript{2+} +
2e\textsuperscript{$-$} $\to$ Sn''. The solution is labeled ``0.60~M
SnSO\textsubscript{4} (0.50~L)''. Orange ovals represent
Sn\textsuperscript{2+} ions and purple ovals represent
SO\textsubscript{4}\textsuperscript{2$-$} ions. An arrow shows
Sn\textsuperscript{2+} migrating toward the cathode. A golden deposit
is visible on the cathode surface. Time is given as 30.0~min.
\end{quote}

\noindent\textbf{Pass~2 (Reasoning):} Receives \textit{only} the above
description and the original question text (no image).

\medskip
\noindent\textbf{What this mode tests.}
For chemistry, Pass~1 must extract precise numerical values and
chemical notation from the diagram. A common failure is misreading
concentrations or charges---e.g., describing
``Sn\textsuperscript{2+}'' as ``Sn\textsuperscript{+}'' would change
the n-factor from 2 to 1, doubling the moles reduced and yielding an
incorrect final concentration of 0.428~M instead of 0.514~M. The
Perception Fidelity gap ($O_H - O_M$) captures exactly these
numerical extraction errors, which are particularly consequential in
quantitative chemistry where small perceptual mistakes cascade into
large calculation errors.

\subsection{Cross-Subject Comparison}
\label{sec:sample_comparison}

Table~\ref{tab:combined_modes} summarizes the diagnostic signals
across both subjects.

\begin{table*}[t]
\centering
\caption{Diagnostic signals for each mode across Biology and Chemistry
examples. The same five-mode framework reveals different failure
patterns depending on the subject and question type.}
\label{tab:combined_modes}
\small
\begin{tabular}{@{}cp{5.5cm}p{5.5cm}@{}}
\toprule
\textbf{Mode} & \textbf{Biology (Reproductive System)}
  & \textbf{Chemistry (Chem Q30: Electrolysis)} \\
\midrule
1 (V+T)
  & Requires identifying labeled structures (A, B, C) + interpreting follicle development cycle + reproductive biology reasoning
  & Requires reading cell setup + Faraday's law calculation \\
\addlinespace
2 (T)
  & Mixed dependency: (a) and (c) answerable from knowledge; (b) unanswerable---``part~B'' is meaningless without the diagram
  & Largely answerable: all numerical data is in the text \\
\addlinespace
3 (V)
  & Dense diagram labels (``Ovary,'' ``Uterus,'' ``Corpus Luteum'') overlap with question text; model must parse which is which
  & Numerical OCR errors (misreading 0.60 as 0.80) cascade into incorrect calculations \\
\addlinespace
4 ($O_H$)
  & Annotations map B~=~Uterus, resolving the critical label; remaining task is physiological reasoning about menstruation
  & Annotations provide n-factor and pre-computed moles; remaining task is arithmetic \\
\addlinespace
5 ($O_M$)
  & Perceptual risk: correctly reading labels ``Ovary'' and ``Uterus'' but failing to associate them with letter labels A, B, C
  & Perceptual risk: misreading charges or concentrations; small errors cause large calculation errors \\
\bottomrule
\end{tabular}
\end{table*}

\noindent The cross-subject comparison reveals a key difference in how
perceptual failures manifest. In Biology, perception failures tend to
be \textit{categorical}---misidentifying a structure leads to a
qualitatively wrong answer. In Chemistry, perception failures tend to
be \textit{quantitative}---misreading a number or charge leads to a
numerically wrong answer through otherwise correct reasoning. Both
failure types are invisible to standard V+T evaluation but are
decomposed by DISSECT's five-mode framework.

\section{Human Oracle Annotation Guidelines}
\label{sec:oracle_guidelines}

Human annotators followed a structured protocol to construct oracle
annotations:

\begin{enumerate}
  \item \textbf{Structural identification:} Label all discrete visual
        entities (molecules, organelles, apparatus components) with their
        standard scientific names.
  \item \textbf{Quantitative extraction:} Transcribe all numerical values,
        measurements, and symbolic notation visible in the image.
  \item \textbf{Spatial relationships:} Describe relative positions,
        connections, and directional indicators (arrows, flow lines).
  \item \textbf{Implicit properties:} Annotate properties that require
        visual interpretation but not domain reasoning (e.g., ``lines
        indicate double bond,'' ``blue shading indicates deoxygenated
        blood'').
  \item \textbf{Question-relevance filtering:} Prioritize annotations
        relevant to the question, but include all identifiable visual
        content to avoid introducing annotator bias about the solution
        path.
\end{enumerate}

\noindent Annotators were undergraduate and graduate students in Chemistry
and Biology. Each annotation was independently verified by a second
annotator, with disagreements resolved by a subject-matter expert.

\end{document}